\documentclass[runningheads]{llncs}

\usepackage{eccv}

\usepackage{eccvabbrv}

\usepackage[accsupp]{axessibility}  % Improves PDF readability for those with disabilities.

\usepackage[pagebackref,breaklinks,colorlinks,citecolor=eccvblue]{hyperref}
\usepackage{etoc}
\usepackage{hyperref}
\usepackage{url}
\usepackage{multirow}
\usepackage{ragged2e}
\usepackage{booktabs}
\usepackage{graphicx}
\usepackage{wrapfig}
\usepackage{comment}
\usepackage{amsmath}
\usepackage[bb=dsserif]{mathalpha}
\usepackage{bm}
\usepackage[dvipsnames]{xcolor}
\usepackage{pifont}
\usepackage{colortbl}
\usepackage{upgreek}
\usepackage{caption}
\usepackage{wrapfig}
\usepackage{tabularx}
\usepackage[all]{hypcap}
\usepackage{orcidlink}
\title{ACE-LoRA: Graph-Attentive Context Enhancement for Parameter-Efficient Adaptation of Medical Vision-Language Models}
\titlerunning{ACE-LoRA}

\newcolumntype{C}{>{\centering\arraybackslash}X}

\begin{document}
\captionsetup[table]{skip=4pt}
\setlength{\textfloatsep}{4pt}

\author{M. Arda Aydın\inst{1,2} \and
Melih B. Yilmaz\inst{1,2} \and
Aykut Koç\inst{1,2} \and
Tolga Çukur\inst{1,2}}

\authorrunning{M.A.~Aydın et al.}

\institute{Bilkent University \and
National Magnetic Resonance Research Center (UMRAM) \\
\email{\{arda.aydin, melih.yilmaz, aykut.koc\}@bilkent.edu.tr, cukur@ee.bilkent.edu.tr}}
%\email{cukur@ee.bilkent.edu.tr}
\maketitle
\begin{abstract}
The success of CLIP-like vision-language models (VLMs) on natural images has inspired medical counterparts, yet existing approaches largely fall into two extremes: specialist models trained on single-domain data, which capture domain-specific details but generalize poorly, and generalist medical VLMs trained on multi-domain data, which retain broad semantics but dilute fine-grained diagnostic cues. Bridging this specialization–generalization trade-off remains challenging. To address this problem, we propose ACE-LoRA, a parameter-efficient adaptation framework for generalist medical VLMs that maintains robust zero-shot generalization. ACE-LoRA integrates Low-Rank Adaptation (LoRA) modules into frozen image-text encoders and introduces an Attention-based Context Enhancement Hypergraph Neural Network (ACE-HGNN) module that captures higher-order contextual interactions beyond pairwise similarity to enrich global representations with localized diagnostic cues, addressing a key limitation of prior Parameter-Efficient Fine-Tuning (PEFT) methods that overlook fine-grained details. To further enhance cross-modal alignment, we formulate a label-guided InfoNCE loss to effectively suppress false negatives between semantically related image-text pairs. Despite adding only 0.95M trainable parameters, ACE-LoRA consistently outperforms state-of-the-art medical VLMs and PEFT baselines across zero-shot classification, segmentation, and detection benchmarks spanning multiple domains. Our code is available at \url{https://github.com/icon-lab/ACE-LoRA}.
\vspace{-0.4cm}
\end{abstract}    

\section{Introduction}
Vision-Language Models (VLMs) \cite{radford2021learning, cherti2023reproducible, jia2021scaling} have rapidly advanced in recent years by learning joint image-text representations at scale. In particular, CLIP \cite{radford2021learning} has achieved remarkable performance in zero-shot classification and has been successfully adapted to a variety of downstream computer vision tasks, including object detection \cite{guopen2022, du2022learning, wu2023cora, li2024learning}, semantic segmentation \cite{zhou2022extract, zhou2023zegclip, aydin2025itaclip, wang2024sclip, lan2024clearclip}, and image generation \cite{tao2023galip, crowson2022vqgan}. This success has sparked growing interest in developing medical VLMs, particularly for radiology and pathology, where large volumes of images and reports must be interpreted as part of routine clinical workflows.

%\begin{figure}[t]
%    \centering
%    \includegraphics[width=\linewidth]{figs/fig1.pdf}
%    \caption{\textbf{Zero-shot classification performance of our fine-tuning approach compared to specialist medical VLMs}, plotted against the number of trainable parameters. All models were pre-trained on MIMIC-CXR v2.0.0 \cite{johnson2019mimic} and evaluated on the RSNA Pneumonia \cite{shih2019augmenting} dataset.}
%    \label{fig_1}
%\end{figure}

Early efforts in this area aimed to develop \emph{specialist} medical VLMs trained on modality-specific datasets to capture domain-relevant visual and textual patterns \cite{huang2021gloria, wang2022multi, cheng2023prior}. However, even one of the most widely used radiology datasets, MIMIC-CXR \cite{johnson2019mimic}, contains only about 377K image-report pairs, orders of magnitude fewer than CLIP’s 400M pairs, limiting the generalization capacity of such models beyond their training data. To alleviate this data bottleneck, recent \emph{generalist} medical VLMs, such as BiomedCLIP \cite{zhang2023biomedclip} and BMC-CLIP \cite{lozano2025biomedica}, leverage large-scale multimodal corpora derived from PubMed \cite{roberts2001pubmed}. While these generalist models are semantically broad, they often lose fine-grained anatomical cues crucial for domain-specific evaluations (e.g., subtle opacity variations in chest X-rays) \cite{sadman2025interpreting}. This fundamental trade-off between specialization and generalization continues to limit the ability of existing medical VLMs to perform robustly in zero-shot settings across new datasets and tasks.

To balance specialization and generalization, we propose to adapt a generalist medical VLM to a specific biomedical domain where visual-textual correspondences are inherently more fine-grained. Capturing such diagnostic cues typically requires domain adaptation; however, retraining or fully fine-tuning large models for every dataset or task is computationally burdensome and clinically impractical. To address this challenge, we fine-tune a generalist foundation model, such as BiomedCLIP \cite{zhang2023biomedclip}, on a paired image-report dataset representative of the domain (e.g., MIMIC-CXR \cite{johnson2019mimic}) to learn domain-specific visual-textual priors. The adapted model can then be transferred to unseen datasets for zero-shot classification without additional dataset-specific training.

A natural approach for efficient model adaptation is Parameter-Efficient Fine-Tuning (PEFT), which updates a small subset of parameters while keeping the pre-trained backbone frozen. Compared to full fine-tuning, PEFT methods are more practical and less prone to overfitting, particularly in data-scarce medical settings \cite{lester2021power}. However, techniques originally developed for natural-image VLMs \cite{zhou2022learning, khattak2023maple, gao2024clip, zanella2024low} face two key limitations in medical imaging. First, they primarily capture global contextual features while overlooking localized patterns that are critical for diagnostic evaluation. Second, most PEFT approaches are designed for few-shot settings that rely on explicitly labeled samples, whereas large-scale medical datasets typically provide image-report pairs rather than curated task labels. This reliance on labeled supervision limits their applicability in low-annotation settings and restricts generalization to unseen datasets.

To address these limitations, we introduce \textbf{ACE-LoRA}, a parameter-efficient framework that enhances generalist medical VLMs by combining Low-Rank Adaptation (\textbf{LoRA}) \cite{hu2022lora} with a novel \textbf{ACE}-HGNN (\textbf{A}ttention-based \textbf{C}ontext \textbf{E}nhancement HGNN) module. Our approach bridges the gap between the domain\allowbreak-specific expertise of specialist models and the strong generalization capability of generalist models through efficient adaptation. Built on BiomedCLIP \cite{zhang2023biomedclip}, ACE-LoRA inserts LoRA modules into both image and text encoders while keeping the backbone frozen. ACE-HGNN then integrates local and global embeddings within each encoder through hypergraph message passing, where hyperedges are constructed from transformer-derived attention affinities and token similarity, enabling structured interactions among groups of semantically related tokens beyond pairwise attention. This allows the model to capture higher-order dependencies among groups of image regions and textual tokens, strengthening cross-modal alignment. In addition, we introduce a label-guided InfoNCE loss that mitigates the false-negative issue common in medical contrastive learning, where distinct samples may share identical clinical semantics. Despite introducing only $\sim$0.95M trainable parameters (approximately 0.48\% of those required for full fine-tuning), ACE-LoRA achieves superior zero-shot performance compared to medical VLMs trained from scratch, highlighting the potential of parameter-efficient adaptation of foundation models for medical imaging. Our main contributions are summarized as follows:

\begin{itemize}

%\vspace{-0.1cm}

\item We demonstrate that parameter-efficient adaptation of a generalist medical VLM enables robust zero-shot transfer across unseen datasets while requiring only minimal additional computation.

\item We propose hypergraph-based context enhancement to model higher-order interactions among global and local embeddings for improved image-text alignment, and introduce a label-guided InfoNCE loss to mitigate false negatives in medical contrastive learning.

\item We benchmark ACE-LoRA against state-of-the-art medical VLMs and PEFT approaches on zero-shot classification, segmentation, and detection tasks.

\vspace{-0.25cm}

\end{itemize}
\section{Related Work}

\subsection{Medical Vision-Language Pretraining}

Recently, medical vision-language pretraining \cite{radford2021learning,bannur2023learning,lai2024carzero, li2024mlip,zhou2023advancing, zhang2023knowledge, wu2023medklip, ozturk2025meta, wang2022medclip, cheng2023prior, zhang2025medunifier, ikezogwo2023quilt, huang2023visual} has emerged as a prominent research area. Existing medical VLMs can be categorized into \textbf{specialist} and \textbf{generalist} models. \textbf{Specialist} medical VLMs are trained on modality-specific datasets containing a relatively limited number of image-text pairs. The pioneering work ConVIRT \cite{zhang2022contrastive} employs a contrastive learning approach that encourages the model to bring matched radiograph-report pairs closer in the embedding space while simultaneously pushing apart mismatched pairs. GLoRIA \cite{huang2021gloria} jointly learns global and local features by aligning image sub-regions with corresponding words. MGCA \cite{wang2022multi} aligns image and text embeddings at region, instance, and disease levels. Yet, specialist models typically fail in generalizing to unseen datasets in the absence of further fine-tuning.  

To enhance model generalization, \textbf{generalist} medical VLMs are trained on broader datasets encompassing multiple imaging modalities. PMC-CLIP \cite{lin2023pmc} builds the multimodal PMC-OA dataset, comprising 1.6M image–text pairs collected from PubMed Central’s Open Access repository. BiomedCLIP \cite{zhang2023biomedclip} curates PMC-15M, a dataset containing 15M image–text pairs, and adapts the original CLIP model to the medical domain by extending the context length. BMC-CLIP \cite{lozano2025biomedica} introduces BIOMEDICA, a large-scale dataset with 24M image-text pairs. Despite these efforts, generalist models still struggle to capture the fine-grained nuances of images within specific imaging modalities. 

\subsection{Parameter-Efficient Fine-Tuning for VLMs}
Efficient fine-tuning strategies for natural-image VLMs have gained significant attention \cite{yang2024mma, guo2025mmrl, yao2023visual, zhu2023prompt, gao2024clip, zhou2022learning, khattak2023maple}, largely due to the substantial computational demands and increased risk of overfitting associated with full fine-tuning. Most current PEFT methods for VLMs employ either \textbf{prompt learning} or \textbf{adapter-based} techniques. In \textbf{prompt learning}, learnable tokens are added to the input or inserted at intermediate layers to better adapt the pre-trained encoders. For instance, CoOp \cite{zhou2022learning} learns optimized tokens that are fed into the text encoder, while CoCoOp \cite{zhou2022conditional} extends CoOp by conditioning the prompts on the corresponding image features to improve generalization. 

In contrast, \textbf{adapter-based} methods use lightweight trainable modules within or on top of frozen encoders. CLIP-Adapter \cite{gao2024clip} appends small residual-style adapters to image-text encoders, while TaskRes \cite{yu2023task} adds a learnable bias to the original text features from the frozen text encoder. Distinct from prompt-learning and adapter-based approaches, CLIP-LoRA \cite{zanella2024low} integrates LoRA \cite{hu2022lora} modules into the query, key, and value projection matrices of both encoders across all layers, resulting in superior performance.

Despite these advances, PEFT strategies for medical VLMs remain underexplored. A recent study, BiomedCoOp \cite{koleilat2025biomedcoop}, utilizes a large language model (LLM) to generate ensemble descriptions for class labels and aligns these descriptions with learnable context tokens to adapt BiomedCLIP. However, this approach, like many PEFT frameworks, is designed for few-shot classification and relies on labeled medical data. In contrast, large-scale medical datasets predominantly provide image-report pairs rather than curated task labels, limiting the applicability of label-dependent adaptation strategies. Furthermore, existing PEFT methods primarily operate on global representations and do not explicitly model the structured relationships among local visual and textual tokens that often convey diagnostic cues. While graph-based modeling has recently gained traction for capturing such structured dependencies, most formulations rely on pairwise interactions between nodes \cite{kipf2017semisupervised, brodyattentive, velivckovic2018graph}. ACE-LoRA addresses this limitation by integrating both local and global embeddings and modeling their contextual relationships through the ACE-HGNN module, which captures higher-order interactions among tokens while retaining parameter-efficient adaptation.

\vspace{-0.3cm}
\section{Method}
\begin{figure*}[t]
    \centering
    \includegraphics[width=\linewidth]{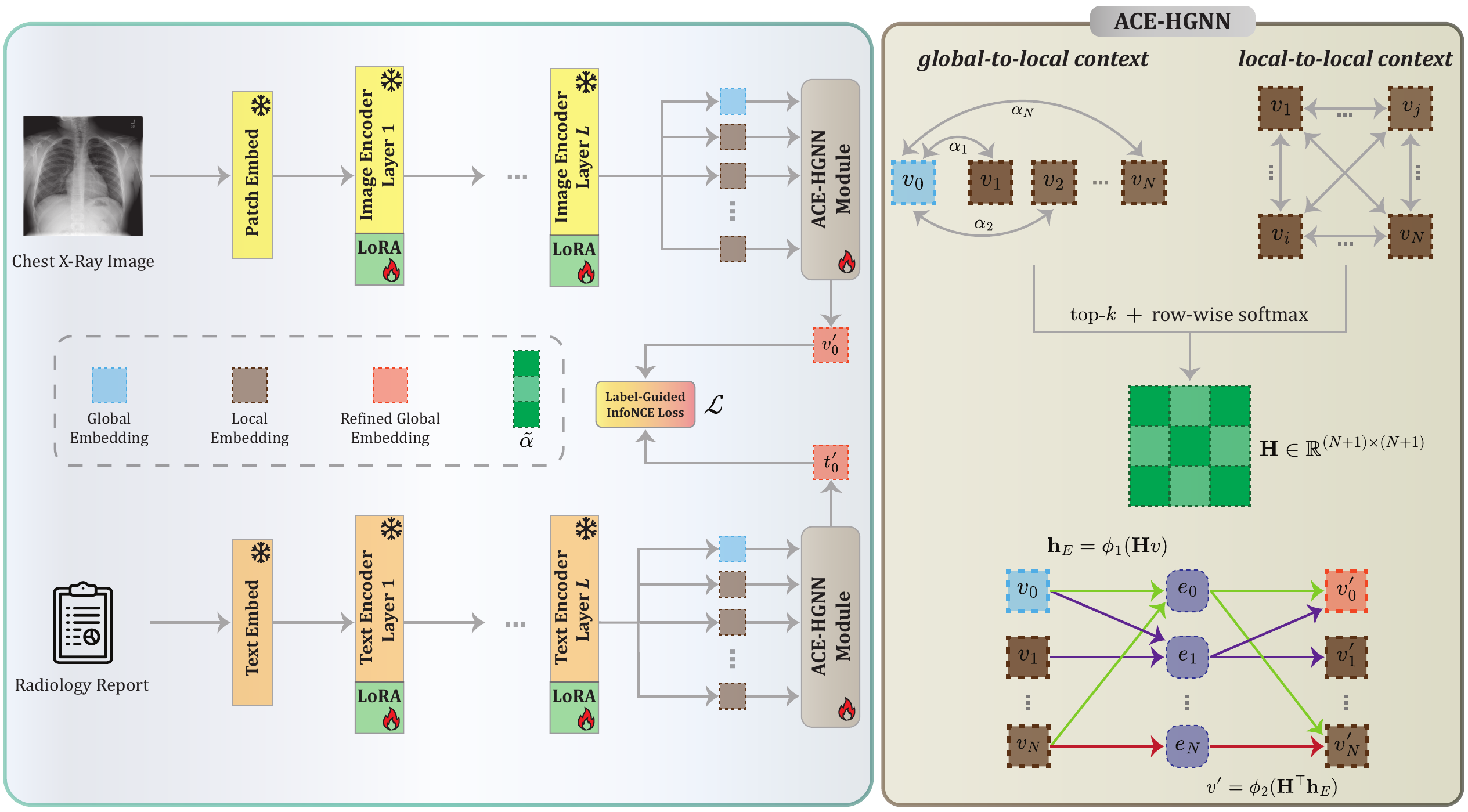}
    \caption{\textbf{Overview of ACE-LoRA.} ACE-LoRA integrates low-rank adaptation modules into self-attention blocks of image and text encoders in a generalist medical VLM, and introduces ACE-HGNN, a hypergraph-based module that models high-order topological dependencies between local (e.g., image patches or report snippets) and global embeddings. For clarity, ACE-HGNN is described using image embeddings, though the same procedure is applied to text embeddings.}
    \label{fig:overview}
\end{figure*}

\vspace{-0.2cm}
The framework of ACE-LoRA is illustrated in Figure~\ref{fig:overview}. First, we integrate LoRA modules into the projection matrices of self-attention layers while keeping the image-text encoders of BiomedCLIP frozen (\S\ref{lora_module}). To capture higher-order structural dependencies in medical images, we introduce the ACE-HGNN module, which models structured token relationships through hypergraph construction (\S\ref{ace-hgnn_module}). Finally, we introduce a label-guided InfoNCE loss to mitigate the false-negative issue in contrastive learning (\S\ref{loss}).

\vspace{-0.3cm}
\subsection{Preliminaries}

The ViT-based \cite{dosovitskiy2020image} CLIP variant employs transformer encoders \cite{vaswani2017attention} for both image and text modalities. The image encoder processes $N$ non-overlapping patches with a prepended \texttt{[CLS]} token. For an input $z^{(l-1)} \in \mathbb{R}^{(N + 1) \times d}$ at layer $l$, the update is formulated as:
\begin{equation}
z' = z^{(l-1)} + \text{SA}(\text{LN}(z^{(l-1)})), \quad z^{(l)} = z' + \text{MLP}(\text{LN}(z')),
\label{eq:transformer_block}
\end{equation}
\noindent where self-attention (SA) is defined by projection matrices $W_{\{Q,K,V,O\}} \in \mathbb{R}^{d \times d}$:
\begin{equation}
\text{SA}(X) = \text{softmax}\left( \frac{XW_Q W_K^T X^T}{\sqrt{d}} \right) X W_V W_O.
\label{eq:attention}
\end{equation}

\noindent The final normalized global image and text embeddings are extracted from the last encoder layers. CLIP is trained to maximize the cosine similarity between matched image-text pairs in a shared latent space. Building on this foundation, BiomedCLIP \cite{zhang2023biomedclip} adapts the original CLIP for medical applications by training on the PMC-15M dataset with extended context length. \emph{Nevertheless, it falls short in capturing domain-specific details within specialized subdomains, underscoring the need for effective fine-tuning to achieve robust domain adaptation.}

\subsection{Integrating LoRA modules}
\label{lora_module}
Motivated by the success of LoRA \cite{hu2022lora} in VLMs for natural images \cite{zanella2024low, kojima2025lorattt}, we integrate LoRA modules into the query ($W_Q$), key ($W_K$), and value ($W_V$) projection matrices within the self-attention modules of both the image and text encoders of BiomedCLIP \cite{zhang2023biomedclip}. Given a pre-trained weight matrix $W_{0} \in \mathbb{R}^{d \times k}$ and an input $x$, the LoRA integration can be expressed as:
\begin{equation}
  h = W_{0}x + \gamma\Delta Wx = W_{0}x + \gamma BAx
  \label{eq:eq_5}
\end{equation}
\noindent where $A \in \mathbb{R}^{r \times k}$ and $B \in \mathbb{R}^{d \times r}$ are low-rank decomposition matrices, $h$ denotes hidden state, $r \ll \min{(d, k)}$ denotes rank, and $\gamma$ is the scaling factor. Following \cite{zanella2024low}, $A$ uses Kaiming initialization \cite{he2015delving} while $B$ is initialized to zero, ensuring that $h$ remains equivalent to $W_{0}x$ at the start of training. Since the original parameters of BiomedCLIP remain frozen and only the decomposition matrices are trainable, injecting LoRA modules introduces minimal memory overhead compared to full fine-tuning.

\subsection{ACE-HGNN Module}
\label{ace-hgnn_module}
While LoRA fine-tuning enhances model adaptability, contrastive training with LoRA modules primarily targets global embeddings, capturing coarse anatomical priors but potentially failing to refine fine-grained cues such as lesion boundaries. \emph{The self-attention mechanism in transformers can effectively model pairwise dependencies among tokens, yet remains limited in capturing higher-order interactions across multiple tokens that characterize localized structures}~\cite{han2022vision, spadaro2025wignet}. \emph{Hypergraph Neural Networks (HGNNs)} \cite{feng2019hypergraph} \emph{offer a natural mechanism to capture such higher-order relations.} 

To leverage these strengths, we draw inspiration from UniGNN \cite{huang2021unignn} and introduce \textbf{ACE-HGNN} (\textbf{A}ttention-based \textbf{C}ontext \textbf{E}nhancement HGNN), a single-layer hypergraph module that treats encoder outputs as vertices and constructs hyperedges from transformer-derived token affinities. This module is applied to both the image and text encoders of BiomedCLIP; we focus on describing the image encoder case below for brevity.

\noindent\textbf{Hypergraph Construction.} Let $v \in \mathbb{R}^{(N+1) \times d}$ denote the image encoder output, consisting of one global and $N$ local embeddings. We construct a hypergraph $\mathcal{G} = (\mathcal{V}, \mathcal{E})$, where the vertex set $\mathcal{V} = \{v_i\}_{i=0}^N$ corresponds to the tokens (i.e., global and local embeddings), and $\mathcal{E}$ denotes the hyperedge set. Each hyperedge $e_i \in \mathcal{E}$ captures the contextual neighborhood of the $i$-th token. \emph{This allows each token to aggregate information from a set of semantically related tokens rather than from isolated pairwise connections, enabling the model to capture structured interactions among groups of patches and the global representation.}

To model the relationships between global and local embeddings, we construct a raw affinity matrix $\mathbf{S} \in \mathbb{R}^{(N+1) \times (N+1)}$. The first row of $\mathbf{S}$, which characterizes the global-to-local context, is derived from the transformer's intrinsic attention maps. We first extract head-wise attention maps $\mathbf{A}_h \in \mathbb{R}^{(N+1)\times(N+1)}$ from each head $h$ of the encoder's final transformer block. After normalizing these maps, they are aggregated by averaging over all heads:
\begin{equation}
\mathbf{A} = \frac{1}{H} \sum_{h=1}^{H} \operatorname{Norm}_{L_2}(\mathbf{A}_h),
\label{eq:agg_attn}
\end{equation}
where $H$ is the number of heads. $\alpha_{i} = \mathbf{A}[0,i]$ quantifies the affinity between the global token and local token $i$. We initialize the first row of \textbf{S} as \footnote{Indices are defined starting from zero ($\{0, \dots, N\}$).}:
\begin{equation}
\mathbf{S}[0, i] = \alpha_{i}, \quad i \in \{1,\ldots,N\}. 
\end{equation}

For local-to-local relationships, we measure semantic alignment via cosine similarity between normalized patch features: 
\begin{equation}
\mathbf{S}_{i, j} = \frac{v_i \cdot v_j}{\| v_i\|_2 \| v_j\|_2} \quad \text{for } i, j \in \{1, \dots, N\}
\end{equation}

\noindent where $v_i$ denotes the $i$-th token in the vertex set.

To filter irrelevant noise and focus on the most informative connections, we apply a top-$k$ filtering mechanism based on $\textbf{S}$, where $k$ is a hyperparameter. For each node $i$, we retain only the set of indices $\Omega_i$ corresponding to the top-$k$ values in row $\mathbf{S}_{i, :}$. We then apply a softmax normalization over these selected elements to obtain normalized hyperedge weights, while enforcing strict self-connections to preserve the identity of pre-trained features ($\mathbf{H}_{i, i} = 1$). Consequently, the entries of our incidence matrix $\mathbf{H} \in \mathbb{R}^{(N+1) \times (N+1)}$ are defined as: 
\begin{equation}
    \mathbf{H}_{i, j} = \begin{cases} 1 & \text{if } i = j \\ \frac{\exp(\mathbf{S}_{i, j})}{\sum_{k \in \Omega_i} \exp(\mathbf{S}_{i, k})} & \text{if } j \in \Omega_i \text{ and } i \neq j \\ 0 & \text{otherwise} \end{cases}
\end{equation}

\noindent Finally, to maintain consistency between global-to-local and local-to-global connections, we enforce symmetry by setting $\textbf{H}[i, 0] = \textbf{H}[0, i]$. 

\noindent \textbf{Hypergraph Message Passing.} We define the information propagation through the hypergraph using two learnable projection functions, $\phi_1$ and $\phi_2$. Both functions share a bottleneck architecture consisting of a linear projection to a lower dimension $d'$, a non-linear activation, and a projection back to $d$:

\begin{equation}
\phi(\mathbf{z}) = \mathbf{W}_{up} \left( \sigma \left( \mathbf{W}_{down} \, \mathbf{z} \right) \right)
\end{equation}

\noindent where $\sigma(\cdot)$ is the LeakyReLU activation.
The message passing occurs in two stages:

\noindent \textbf{Vertex-to-Hyperedge.} We first aggregate information from the nodes to construct hyperedge features. This step captures the contextual information defined by the incidence matrix $\mathbf{H}$:
\begin{equation}   
\mathbf{h}_{E} = \phi_1(\mathbf{H} v)
\end{equation}

\noindent \textbf{Hyperedge-to-Vertex.} Finally, we map the hyperedge features back to the vertex domain, allowing nodes to incorporate higher-order feedback from the hyperedges they influence:
\begin{equation}
    v' = \phi_2(\mathbf{H}^\top \mathbf{h}_{E})
\end{equation}

\noindent The final output $v'$ serves as the refined feature representation, enriched with both attention-guided context and local patch-similarity structures. Unlike the original UniGAT architecture~\cite{huang2021unignn}, ACE-HGNN does not learn attention coefficients independently; instead, it leverages \emph{transformer-derived affinities} to define the underlying hypergraph topology. By coupling self-attention with hypergraph message passing, our approach preserves transformer-based global context modeling while enabling structured aggregation over groups of semantically related tokens. This formulation enables the model to emphasize informative local regions while maintaining coherent global representations. Furthermore, our HGNN formulation naturally reduces to a standard Graph Neural Network (GNN) when $\phi_{1}(z) = z$; we also evaluate this simplified variant in \S\ref{ablation}.

\begin{figure}[t]
    \centering
    \includegraphics[width=0.65\linewidth]{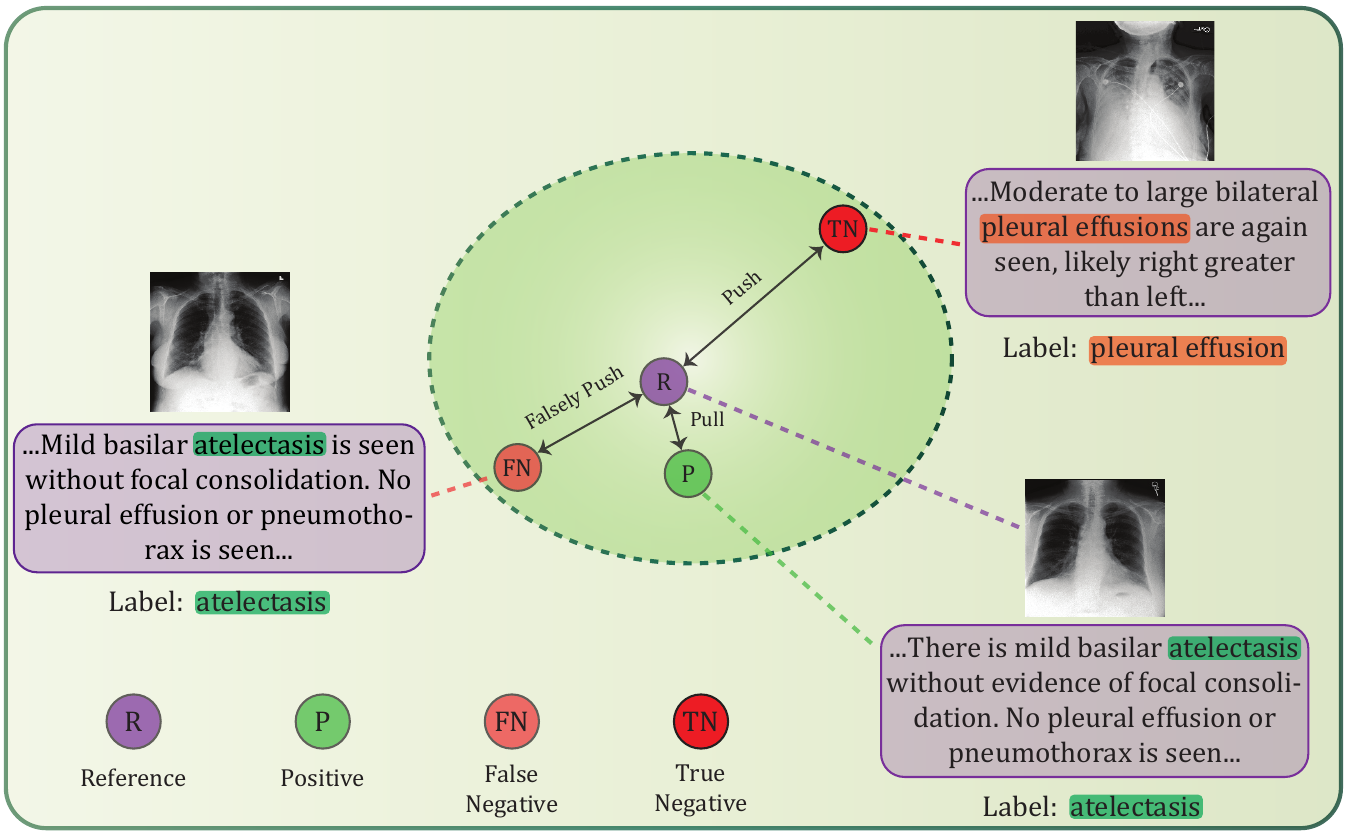}
    \caption{\textbf{False negatives in contrastive learning.} The CLIP loss treats all non-matching pairs as negatives, which can falsely push apart semantically similar samples, whereas our formulation avoids separating pairs that share the same disease label.}
    \label{fig_2}
\end{figure}

\subsection{Label-Guided InfoNCE Loss}
\label{loss}
\emph{CLIP-like VLMs assume that all non-matching image-text pairs are negative samples; however, reports from different images may describe the same disease pathology and are thus incorrectly treated as false negatives} (see Figure \ref{fig_2}), leading to suboptimal results, as discussed in \cite{wang2022medclip}. To alleviate this issue, we incorporate disease labels extracted from each report using the CheXpert \cite{irvin2019chexpert} labeler on the MIMIC-CXR v2.0.0 dataset \cite{johnson2019mimic}. \emph{We then reformulate the InfoNCE loss \cite{oord2018representation} such that, when a non-matching image-text pair shares the same disease label as the reference pair, the model neither attracts nor repels their embeddings but instead excludes the pair from contributing to the loss.} The overall loss over a minibatch of size $B$ is formulated as:

\begin{equation}
\small
\begin{split}
\mathcal{L}
  = -\frac{1}{2B} \Bigg(
     &\sum_{i=1}^{B}
      \log
      \frac{
        \exp\big(\langle v_{0}^{\prime(i)}, t_{0}^{\prime(i)} \rangle / \tau\big)
      }{
        \sum_{k=1}^{B} \mathbb{1}_{i,k}
        \exp\big(\langle v_{0}^{\prime(i)}, t_{0}^{\prime(k)} \rangle / \tau\big)
      }
\\[0.6ex]
   +\;
   &\sum_{i=1}^{B}
      \log
      \frac{
        \exp\big(\langle t_{0}^{\prime(i)}, v_{0}^{\prime(i)} \rangle / \tau\big)
      }{
        \sum_{k=1}^{B} \mathbb{1}_{i,k}
        \exp\big(\langle t_{0}^{\prime(i)}, v_{0}^{\prime(k)} \rangle / \tau\big)
      }
  \Bigg)
\end{split}
\label{eq:eq_9}
\end{equation}

\noindent where $\mathbb{1}_{i,k} \in \{0,1\}$ denotes an indicator function that equals 1 when the disease label of the $i$-th pair differs from that of the $k$-th pair or when $i=k$, and 0 otherwise. $v_{0}^{\prime (i)}$ and $t_{0}^{\prime (i)}$ represent the refined global image and text embeddings of the $i$-th pair, respectively. $\langle ., .\rangle$ denotes cosine similarity, and $\tau$ is a temperature parameter initialized to the pre-trained value from BiomedCLIP and updated during fine-tuning. Note that the refined global embeddings in the above equation are normalized across the embedding dimension, although this normalization is omitted from the notation for simplicity.

%\vspace{3pt}
%\noindent\textbf{Intuition.}
%By masking out pairs that share identical disease labels, the label-guided InfoNCE loss concentrates on truly contrasting samples instead of penalizing semantically aligned image–text pairs. This objective prevents feature collapse and encourages embeddings to align with clinically meaningful rather than superficial visual patterns.

%By masking out pairs that share identical disease labels, the label-guided InfoNCE loss focuses optimization on genuinely contrasting cases rather than penalizing semantically aligned image–text pairs. This objective mitigates representation collapse among disease-relevant features and encourages embeddings to organize along clinically meaningful dimensions instead of superficial visual similarities.

\section{Experiments}

\subsection{Training Setup}

\textbf{Dataset.} Taking BiomedCLIP \cite{zhang2023biomedclip} as our base model, we adapt it on the MIMIC-CXR dataset \cite{johnson2019mimic} for radiology. The dataset comprises 377,110 radiographs from 227,827 imaging studies. Following previous studies \cite{wang2022multi, cheng2023prior}, we exclude lateral images, as our target downstream datasets contain only frontal-view chest radiographs. For histopathology, we construct a dedicated adaptation dataset by extracting histopathology-related images from the PMC-OA \cite{lin2023pmc} corpus. This process yields a refined subset of 160K image-text pairs;  further details on the dataset construction process are provided in the Appendix (\S\ref{dataset_appendix}).

\noindent \textbf{Implementation Details.} Images are resized to $224 \times 224$ to ensure compatibility with the model architecture. We train the model for 40 epochs (15 epochs for histopathology) on the MIMIC-CXR dataset using three RTX 3090 GPUs, with a batch size of 64 per device. We employ the AdamW \cite{loshchilovdecoupled} optimizer with a learning rate of $1\mathrm{e}{-3}$ and a weight decay of $1\mathrm{e}{-2}$. A linear warmup is applied during the first epoch, followed by a cosine annealing learning rate scheduler \cite{loshchilov2017sgdr}. For LoRA, we set the rank parameter $r$ to 4 and the scaling factor $\gamma$ to 1. For the top-$k$ operation, we set $k$ to 5.

\begin{table*}[t]
\centering
\begin{tabular}{l cc|cc|cc}
\toprule
\multirow{2}{*}{\textbf{Method}} 
  & \multicolumn{2}{c|}{\textbf{CheXpert 5$\times$200}} 
  & \multicolumn{2}{c|}{\textbf{RSNA}} 
  & \multicolumn{2}{c}{\textbf{SIIM}} \\
  & ACC (\%) & AUC (\%) 
  & ACC (\%) & AUC (\%)
  & ACC (\%) & AUC (\%) \\
\midrule
BiomedCLIP \cite{zhang2023biomedclip} & 35.50 & 67.46 & 74.34 & 81.14 & 61.40 & 64.31 \\
\midrule
CoOp \cite{zhou2022learning} & 40.90 & 69.32 & 75.07 & 79.92 & 65.69 & 69.87 \\
CoCoOp \cite{zhou2022conditional} & 40.10 & 69.74 & 75.02 & 79.95 & 67.24 & 71.21 \\
CLIP-Adapter \cite{gao2024clip} & 42.70 & 77.55 & 68.47 & 75.14 & 60.38 & 61.26 \\
TaskRes \cite{yu2023task} & 36.50 & 68.17 & 74.77 & 81.36 & 63.53 & 67.14 \\
MaPLe \cite{khattak2023maple} & 45.90 & 77.45 & 73.34 & 83.34 & 71.47 & \underline{80.59} \\
MMA \cite{yang2024mma} & 44.00 & \underline{79.34} & \underline{74.89} & \underline{83.39} & 63.97 & 67.12 \\
CLIP-LoRA \cite{zanella2024low} & 45.80 & 77.51 & 72.85 & 77.74 & \underline{72.50} & 80.35 \\
MMRL \cite{guo2025mmrl} & 45.20 & 75.22 & 71.62 & 80.55 & 68.15 & 76.19 \\
\midrule
Full FT & \underline{47.40} & 71.63 & 73.94 & 77.33 & 70.50 & 72.28 \\
\midrule
\rowcolor{SeaGreen!70} ACE-LoRA & \textbf{49.80} & \textbf{80.87} & \textbf{79.54} & \textbf{87.19} & \textbf{73.35} & \textbf{81.51} \\
\bottomrule
\end{tabular}
\caption{\textbf{Comparison of zero-shot classification performance across different fine-tuning methods.} Full FT represents full fine-tuning, in which all parameters of the pre-trained model are updated. All methods are trained using Label-guided InfoNCE loss to ensure a fair comparison. The best results are highlighted in \textbf{bold}, and the second-best are \underline{underlined}.}
\label{tab:tab_1}
\end{table*}

\subsection{Downstream Tasks and Experimental Setup}

\textbf{Zero-Shot Image Classification.} We evaluate ACE-LoRA on zero-shot image classification to assess its generalization capabilities. In the following experiments, for medical VLMs trained from scratch, we do not perform additional retraining; instead, we utilize the publicly available checkpoints from the official implementations to benchmark their performance on downstream datasets. We compare ACE-LoRA against both state-of-the-art medical VLMs trained from scratch and PEFT methods for VLMs. Accuracy (ACC) and area under the ROC curve (AUC) are used as evaluation metrics across all datasets.

\textbf{a) Radiology.} Similar to CLIP, we construct two prompt templates to guide the mapping from text to image features: \texttt{"a chest X-ray image of \{disease\}"} and \texttt{"Findings \allowbreak suggesting \allowbreak \{disease\}"}. We conduct experiments on three datasets: (1) \textbf{CheXpert 5$\times$200}: consistent with prior studies \cite{huang2021gloria, cheng2023prior}, we use a subset of the CheXpert \cite{irvin2019chexpert} dataset containing 200 images per disease class, with each image annotated with a single disease, (2) \textbf{RSNA Pneumonia} \cite{shih2019augmenting}, which comprises 29,684 frontal-view chest X-ray images, each labeled as either pneumonia or normal. For the classification task, we randomly sample 15\% of the images as the test set, and (3) \textbf{SIIM-ACR Pneumothorax} \cite{SIIM-ACR}, which contains over 12,000 chest radiographs, each categorized as either pneumothorax or normal. To obtain a comparable number of test images to the RSNA dataset, we randomly select 30\% of the SIIM dataset as the test set. We utilize 20\% of the NIH ChestX-ray14 \cite{wang2017chestx} test set for hyperparameter tuning, given its status as a representative and large-scale radiology benchmark.   

\begin{table*}[t]
\centering
\begin{tabular}{l cc|cc|cc}
\toprule
\multirow{2}{*}{\textbf{Method}} 
  & \multicolumn{2}{c|}{\textbf{CheXpert 5$\times$200}} 
  & \multicolumn{2}{c|}{\textbf{RSNA}} 
  & \multicolumn{2}{c}{\textbf{SIIM}} \\
  & ACC (\%) & AUC (\%) 
  & ACC (\%) & AUC (\%)
  & ACC (\%) & AUC (\%) \\
\midrule
\multicolumn{7}{l}{\textit{Specialist Medical VLMs}} \\
\midrule
ConVIRT \cite{zhang2022contrastive} & 45.10 & 75.82 & 42.64 & 42.13 & 46.54 & 49.09 \\
GLoRIA \cite{huang2021gloria} & 24.80 & 71.60 & 68.42 & 76.73 & 48.67 & 50.13 \\
MGCA \cite{wang2022multi} & 21.40 & 52.62 & 63.75 & 68.21 & 58.52 & 62.55 \\
BioVIL \cite{bannur2023learning} & \underline{48.40} & \underline{76.99} & 56.66 & 62.23 & 53.32 & 59.99 \\
PRIOR \cite{cheng2023prior} & 40.10 & 72.68 & 60.33 & 62.71 & 60.38 & 63.25 \\
KAD \cite{zhang2023knowledge} & 23.50 & 51.54 & 70.17 & 66.75 & 57.19 & 51.22 \\
\midrule

\multicolumn{7}{l}{\textit{Generalist Medical VLMs}} \\
\midrule
PMC-CLIP \cite{lin2023pmc} & 23.00 & 52.58 & 60.93 & 64.59 & 51.02 & 51.57 \\
BMC-CLIP \cite{lozano2025biomedica} & 24.70 & 55.99 & 53.83 & 56.18 & 48.67 & 47.07 \\
BiomedCLIP \cite{zhang2023biomedclip} & 35.50 & 67.46 & \underline{74.34} & \underline{81.14} & \underline{61.40} & \underline{64.31} \\
\midrule
\rowcolor{SeaGreen!70} ACE-LoRA & \textbf{49.80} & \textbf{80.87} & \textbf{79.54} & \textbf{87.19} & \textbf{73.35} & \textbf{81.51} \\
\bottomrule
\end{tabular}
\caption{\textbf{Zero-shot classification performance on the CheXpert 5$\times$200, RSNA, and SIIM datasets.} Competing methods are categorized into specialist and generalist medical VLMs.}
\label{tab:tab_2}
\vspace{-0.25cm}
\end{table*}

\begin{table}[h]
\centering

\begin{minipage}[t]{0.48\textwidth}
\vspace{0pt}
\centering
\setlength{\tabcolsep}{2pt}
\begin{tabular}{l c|c|c}
        \toprule
        \textbf{Method} & \textbf{Lung} & \textbf{Colon} & \textbf{MHIST} \\
        \midrule
        \multicolumn{4}{l}{\textit{Specialist Medical VLMs}} \\
        \midrule
        PLIP \cite{huang2023visual}     & 78.77 & 77.79 & 56.63 \\
        QuiltNet \cite{ikezogwo2023quilt}      & \underline{81.87} & 87.10 & \underline{56.82} \\
        \midrule
        \multicolumn{4}{l}{\textit{Generalist Medical VLMs}} \\
        \midrule
        BMC-CLIP \cite{lozano2025biomedica}  & 76.71 & \underline{87.43} & 41.78 \\
        BiomedCLIP \cite{zhang2023biomedclip} & 72.90 & 85.33 & 34.04 \\
        \midrule
        \rowcolor{SeaGreen!70} 
        ACE-LoRA & \textbf{84.03} & \textbf{90.39} & \textbf{57.27} \\
        
        \bottomrule
    \end{tabular}

    \caption{\textbf{Zero-shot classification accuracy on histopathology datasets.} ``Lung'' and ``Colon'' denote the respective subsets of the LC25000 dataset.}
\label{tab:tab_patho}
\end{minipage}
\hfill
\begin{minipage}[t]{0.48\textwidth}
\vspace{0pt}
\centering
\begin{tabular}{l c | c}
\toprule
\multirow{2}{*}{\textbf{Method}} &  \multicolumn{1}{c}{\textbf{SIIM}} 
& \multicolumn{1}{c}{\textbf{RSNA}} \\
& \textbf{Dice (\%)} 
& \textbf{mAP (\%)} \\
\midrule
GLoRIA \cite{huang2021gloria}           & 40.02  & \underline{20.42} \\
MGCA \cite{wang2022multi}           & 44.53  & 20.30  \\
BiomedCLIP \cite{zhang2023biomedclip}     & 44.63  & 20.25  \\
KAD \cite{zhang2023knowledge}           & 45.17  & 18.10 \\ 
BioVIL \cite{bannur2023learning}        & 45.73  & 19.31  \\
PRIOR \cite{cheng2023prior}         & \underline{45.85}  & 18.89  \\
\midrule
\rowcolor{SeaGreen!70} ACE-LoRA            & \textbf{46.34}  & \textbf{21.29} \\
\bottomrule
\end{tabular}
\caption{\textbf{Semantic segmentation and object detection results on the SIIM and RSNA datasets.} Dice scores are reported for the \textit{Pneumothorax} class.}
\label{tab:tab_3}
\end{minipage}
\vspace{-0.25cm}
\end{table}

\textbf{b) Histopathology.} To demonstrate generalizability beyond radiology, we evaluate our method on histopathology benchmarks. For fair comparison, we adopt the prompt templates from the QuiltNet \cite{ikezogwo2023quilt} method. Experiments are conducted on three datasets: (1) \textbf{LC25000 (Lung)} and (2) \textbf{LC25000 (Colon)}, subsets of LC25000 \cite{borkowski2019lung} containing 15,000 and 10,000 images of lung and colon adenocarcinomas, respectively (with 5,000 images per class), and (3) \textbf{MHIST} \cite{wei2021petri}, a binary classification dataset consisting of 3,152 images of colorectal polyps. For validation, we use 10\% of the filtered PMC-OA dataset. We employ the original CLIP loss during training, as the inherent complexity of histopathology reports and the absence of a standardized labeler, analogous to CheXpert for radiology, limit the applicability of automated label extraction. Additional details about the datasets and the exact prompts are provided in the Appendix (\S\ref{prompt_templates}).

\noindent \textbf{Semantic Segmentation.} To maintain compatibility with different image encoder architectures used in medical VLMs (CNNs and ViTs), we employ UPerNet \cite{xiao2018unified} as the segmentation decoder across all methods, following \cite{zhou2024benchx}. We keep the pre-trained image encoder frozen as the backbone while fine-tuning only the UperNet decoder parameters. We use images containing pneumothorax from the SIIM \cite{SIIM-ACR} dataset and split the dataset into 70\%/15\%/15\% for training, validation, and testing, respectively. The Dice score is used as the evaluation metric.

\noindent \textbf{Object Detection.} Following \cite{wang2022multi}, we use YOLOv3 \cite{redmon2018yolov3} as the detection network, keeping the pre-trained image encoder frozen as the backbone. As in the segmentation task, RSNA \cite{shih2019augmenting} images annotated with pneumonia are analyzed, and the dataset is split into 70\%/15\%/15\% for training, validation, and testing, respectively. We evaluate performance using mean average precision (mAP), calculated across IoU thresholds of 0.4, 0.45, 0.5, 0.55, 0.6, 0.65, 0.7, and 0.75.
\vspace{-0.35cm}
\subsection{Results}
\textbf{Results on Zero-Shot Image Classification.} To demonstrate ACE-LoRA over existing PEFT methods, we conduct a comparative analysis against established approaches. As listed in Table \ref{tab:tab_1}, our method surpasses other PEFT strategies in zero-shot classification across all datasets. Among competing methods, multimodal PEFT methods (e.g., MaPLe \cite{khattak2023maple}, MMA \cite{yang2024mma}) generally outperform unimodal approaches (e.g., CoOp \cite{zhou2022conditional}), suggesting that cross-modal alignment benefits medical VLMs. While full fine-tuning updates all $\sim$197M parameters of the pre-trained bacbkbone, ACE-LoRA achieves superior performance with only $\sim$0.95M trainable parameters, demonstrating superior parameter efficiency.

To further benchmark ACE-LoRA, we evaluate our method against state-of-the-art medical VLMs on the zero-shot classification task, with results reported in Table \ref{tab:tab_2}. Our method achieves notable performance improvements over both specialist and generalist medical VLMs on the CheXpert 5$\times$200, RSNA, and SIIM datasets. Consistent with prior studies \cite{li2024mlip,cheng2023prior}, we observe that specialist medical VLMs \cite{wang2022multi,huang2021gloria, zhang2022contrastive,bannur2023learning, cheng2023prior,zhang2023knowledge} frequently fail to generalize beyond their pre-training datasets when evaluated in zero-shot settings using simple prompt templates. Although some methods achieve strong performance on a single dataset, they struggle to maintain consistent results across different datasets. Furthermore, we find that generalist medical VLMs \cite{zhang2023biomedclip,lozano2025biomedica, lin2023pmc} tend to underperform on chest X-ray datasets, likely due to the domain-specific characteristics of radiographs. Note that many PEFT methods yield more competitive results than VLMs trained from scratch, highlighting the efficacy of fine-tuning approaches.

Finally, we evaluate our framework on histopathology benchmarks, comparing it against both pathology-specialized and generalist medical VLMs (Table \ref{tab:tab_patho}). Our approach yields substantial improvements over the BiomedCLIP baseline across three datasets and even outperforms specialized models such as PLIP \cite{huang2023visual} and QuiltNet \cite{ikezogwo2023quilt}. Notably, ACE-LoRA achieves these results while using significantly fewer image-text pairs (140K) during training compared to PLIP (210K) and QuiltNet (1M), demonstrating superior data efficiency.

\begin{table}[t]
\centering
\begin{minipage}[t]{0.65\textwidth}
%\tablestyle{-13pt}{1.1}
\arrayrulecolor{black}
%\setlength\arrayrulewidth{1pt}
%\addtolength{\tabcolsep}{+16pt}
\resizebox{\columnwidth}{!}{%
\begin{tabular}{ccc ccc ccccc }
\toprule
\multicolumn{3}{c}{\textbf{Components}}     & \multicolumn{3}{c}{\textbf{Datasets}}                                          \\ \midrule
 \textbf{LoRA} & \textbf{ACE-HGNN}                   & \textbf{Label}                   & \textbf{\textbf{CheXpert}}           & \textbf{RSNA}          & \textbf{SIIM}         \\ \midrule
\ding{55} & \ding{55} & \ding{55} & 35.50          & 74.34          & 61.40      \\
\ding{51} & \ding{55} & \ding{55} & 45.60          & 67.07         & 68.29          \\
\ding{51} & \ding{51} & \ding{55} & \underline{49.20}          & \underline{74.64}          & 70.56      \\
\ding{51} & \ding{55} & \ding{51} & 45.80    & 72.85  &  \underline{72.50}          \\
\rowcolor{SeaGreen!70}\ding{51} & \ding{51} & \ding{51} & \textbf{49.80} & \textbf{79.54} & \textbf{73.35} \\ \bottomrule
\end{tabular}}
  \caption{\textbf{Impact of each component in ACE-LoRA on classification accuracy.} ``Label'' denotes the Label-guided InfoNCE loss. When this loss is not applied (\ding{55}), the CLIP loss is used during training.}
  \label{tab:tab_4}
\end{minipage}
%\vspace{-0.5cm}
\end{table}

\noindent \textbf{Results on Semantic Segmentation.} We evaluate our approach against existing methods on the semantic segmentation task using the SIIM dataset to assess its dense prediction performance (Table \ref{tab:tab_3}). Although the performance gap among methods is more moderate than the previous task, ACE-LoRA still remains the top performer, with its design sensitive to local context and high-order topological interactions substantially improving the base model, BiomedCLIP.

\noindent \textbf{Results on Object Detection.} Table \ref{tab:tab_3} also reports the object detection performance of medical VLMs on the RSNA dataset. ACE-LoRA outperforms competing medical VLMs in detecting pneumonia, demonstrating its ability to capture subtle pathological patterns in radiographs. Notably, models that emphasize local alignment (e.g., GLoRIA \cite{huang2021gloria} and MGCA \cite{wang2022multi}) yield second-best results after ACE-LoRA, underscoring the importance of local information.

\subsection{Ablation Studies}
\label{ablation}

\textbf{Impact of method components.} We analyze the contribution of each component in Table \ref{tab:tab_4}. The first row corresponds to the pre-trained BiomedCLIP model, which serves as our baseline. Injecting LoRA modules improves overall performance, particularly on the CheXpert 5$\times$200 dataset, although it leads to a slight performance degradation on the RSNA dataset. Incorporating the ACE-HGNN module further enhances the performance on the CheXpert 5$\times$200 and SIIM datasets, while restoring performance on RSNA. Notably, ACE-LoRA outperforms all medical VLMs trained from scratch even without the Label-guided InfoNCE loss, demonstrating that our method achieves strong performance using the original CLIP loss alone. Finally, the combination of all modules with the Label-guided InfoNCE loss achieves the best overall results, significantly surpassing the initial baseline. 

%We view LoRA and ACE-HGNN as complementary: LoRA facilitates adaptation across all layers, whereas ACE-HGNN enhances local context awareness.

\begin{table}[t]
\centering
\setlength{\tabcolsep}{1pt}
\begin{minipage}[t]{0.48\textwidth}
\captionsetup{width=\linewidth}
\vspace{0pt}
\begin{tabularx}{\linewidth}{l 
    >{\hsize=1.4\hsize}C 
    >{\hsize=0.8\hsize}C 
    >{\hsize=0.8\hsize}C}
\toprule
{\textbf{Method}} & \textbf{CheXpert} & \textbf{RSNA} & \textbf{SIIM} \\
\midrule
GAT          & \underline{46.90} & \underline{76.27} & \underline{71.22} \\
GATv2        & \underline{46.90} & 75.89 & 71.08 \\
\rowcolor{SeaGreen!70} Ours             & \textbf{49.80} & \textbf{79.54} & \textbf{73.35}\\
\bottomrule
\end{tabularx}
\caption{\textbf{Effect of using attention maps as edge weights.} Detailed formulations of GAT \cite{velivckovic2018graph} and GATv2 \cite{brodyattentive} are provided in the Appendix (\S\ref{gat_gatv2_formula}).}
\label{tab:tab_6}
\end{minipage}
\hfill
\begin{minipage}[t]{0.48\textwidth}
\captionsetup{width=\linewidth}
\vspace{0pt}
\begin{tabularx}{\linewidth}{l 
    >{\hsize=1.4\hsize}C 
    >{\hsize=0.8\hsize}C 
    >{\hsize=0.8\hsize}C}
\toprule
\textbf{Method} & \textbf{CheXpert} & \textbf{RSNA} & \textbf{SIIM} \\
\midrule
GNN   & 48.10 & 76.94 & 69.40 \\
\rowcolor{SeaGreen!70}
HGNN & \textbf{49.80} & \textbf{79.54} & \textbf{73.35} \\
\bottomrule
\end{tabularx}
\caption{\textbf{Importance of the hypergraph formulation.} Zero-shot accuracy results are reported.}
\label{tab:hyper}
\end{minipage}
%\vspace{-0.25cm}
\end{table}

\noindent \textbf{Importance of attention-driven hyperedge weights.} We conduct an ablation study to evaluate the effectiveness of using attention maps as hyperedge representations (Table \ref{tab:tab_6}). We compare our approach with alternative designs that learn attention coefficients from scratch, adopting the attention formulations of GAT \cite{velivckovic2018graph} and GATv2 \cite{brodyattentive}. For a fair comparison, the overall architecture is kept identical, and only the mechanism for computing attention scores in the constructed hypergraph is modified. Results demonstrate that using attention maps as hyperedge representations leads to improved performance, thereby validating the proposed design.

\noindent \textbf{Importance of hypergraph formulation.} To assess the importance of capturing high-order interactions among groups of tokens via HGNNs compared to pairwise interactions in GNNs, we benchmark a variant where $\phi_{1}(z) = z$ (Table \ref{tab:hyper}). This constraint reduces our hypergraph formulation to a conventional GNN. We observe that the HGNN variant consistently achieves superior results across all three benchmarks, confirming that higher-order relationship modeling is essential for capturing the complex dependencies in medical images.

\begin{table}[t]
\centering
\setlength{\tabcolsep}{4pt}
\begin{tabular}{l ccc}
\toprule
{\textbf{Method}} & \textbf{CheXpert} & \textbf{RSNA} & \textbf{SIIM} \\
\midrule
BMC-CLIP    & 24.70 & 53.83 & 48.67 \\
\rowcolor{SeaGreen!70} \textbf{+} ACE-LoRA & \textbf{50.50}  & \textbf{67.35} & \textbf{70.70} \\
$\Updelta$  & \textcolor{green!40!black}{\textbf{+25.80}} & \textcolor{green!40!black}{\textbf{+13.52}} & \textcolor{green!40!black}{\textbf{+22.03}} \\
\bottomrule
\end{tabular}
\caption{\textbf{Performance of ACE-LoRA when deployed on an alternative generalist medical VLM (BMC-CLIP).} $\Updelta$ denotes the performance boost in zero-shot accuracy that ACE-LoRA provides when adopted for BMC-CLIP.}
\label{tab:bmc_clip}
%\vspace{-6pt}
\end{table}

\noindent \textbf{Generalization to alternative generalist VLMs.} To analyze the performance of our method when integrated with alternative generalist medical VLMs, we apply our framework to BMC-CLIP. The results, reported in Table \ref{tab:bmc_clip}, show that ACE-LoRA consistently improves zero-shot performance across all three benchmarks. Notably, the performance gains observed with BMC-CLIP are highly competitive with those achieved using BiomedCLIP, indicating that our approach remains robust and flexible across different medical VLM backbones.

%\begin{wraptable}{r}{0.5\textwidth}
%\vspace{-0pt}
%\centering
%\begin{tabular}{l ccc}
%\toprule
%\textbf{Method} & \textbf{ CheXpert} & \textbf{ RSNA} & \textbf{ SIIM} \\
%\midrule
%GNN   & 48.10 & 76.94 & 69.40 \\
%\rowcolor{SeaGreen!70}
%HGNN & \textbf{49.80} & \textbf{79.54} & \textbf{73.35} \\
%\bottomrule
%\end{tabular}
%\caption{\textbf{Effect of using hypergraphs.}}
%\label{tab:hyper}
%\end{wraptable}

\begin{figure}[t]
%\vspace{-0.65cm}
    \centering
    \includegraphics[width=\linewidth]{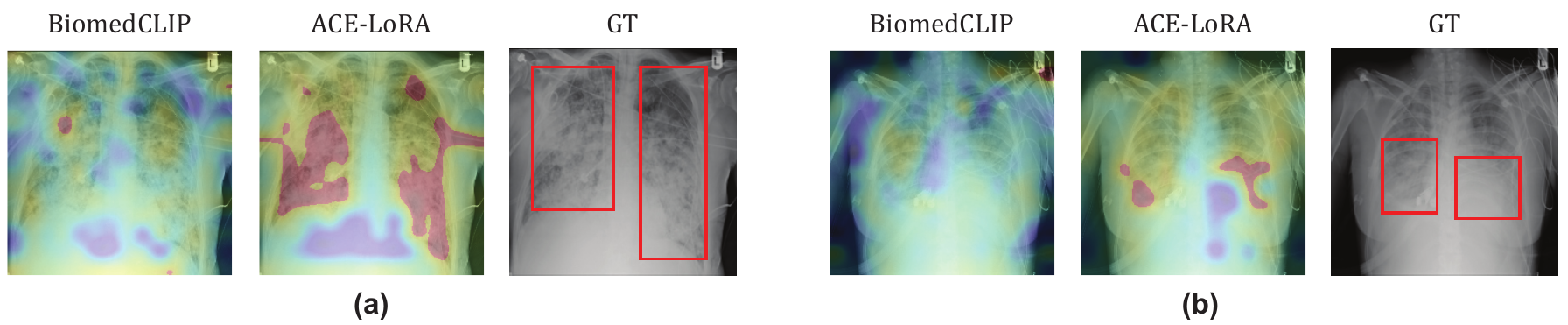}
    %\vspace{-0.45cm}
    \caption{\textbf{Comparison of cross-modal similarity maps on the RSNA Dataset.} Similarity maps show the correspondence between image regions and the text query \textit{``Pneumonia''}, with red boxes indicating ground-truth (GT) abnormal regions.}
    %\textbf{(a)} In this case, the abnormalities are distributed across both lungs. ACE-LoRA successfully highlights these regions, whereas BiomedCLIP fails to attend to them. \textbf{(b)} Here, both lungs contain localized pathological regions. While BiomedCLIP detects only one region, ACE-LoRA accurately localizes both.}
    \label{fig_4}
\end{figure}
\vspace{0.5cm}

\noindent \textbf{Visualization of cross-modal similarity maps.} 
To assess the localization capabilities of our method, we evaluate the visual similarity maps generated by ACE-LoRA and BiomedCLIP against ground-truth detection labels from the RSNA dataset. As illustrated in Figure \ref{fig_4}(a), ACE-LoRA successfully highlights abnormal regions in a case of distributed pathology, whereas BiomedCLIP fails to identify the relevant findings. Similarly, in a case involving localized bilateral pathology (Figure \ref{fig_4}(b)), ACE-LoRA accurately localizes both abnormalities, while BiomedCLIP is unable to focus on pathological regions of interest.

\section{Conclusion}
In this study, we present ACE-LoRA, an efficient fine-tuning strategy that adapts a generalist medical VLM for robust zero-shot transfer to unseen domain-specific datasets. By refining both local and global embeddings, ACE-LoRA more effectively captures the semantic complexity of medical images. Our results show that this lightweight, locality-aware approach enables BiomedCLIP to achieve strong performance in data-limited clinical settings. Moreover, the strong zero-shot generalization of PEFT-based methods over medical VLMs trained from scratch underscores the promise of this research direction.

\bibliographystyle{splncs04}
\bibliography{main}

\clearpage
\newpage
\appendix
\makeatletter
\renewcommand{\theHsection}{appendix\thesection}
\renewcommand{\theHsubsection}{appendix\thesubsection}
\makeatother

\renewcommand{\thetable}{A.\arabic{table}}
\renewcommand{\thefigure}{A.\arabic{figure}}
\setcounter{table}{0}
\setcounter{figure}{0}

\begin{center}
    {\Large \textbf{Appendix}}
\end{center}

% Reset figure/table numbering
\renewcommand{\thetable}{A.\arabic{table}}
\renewcommand{\thefigure}{A.\arabic{figure}}
\setcounter{table}{0}
\setcounter{figure}{0}

\textbf{Table of Contents}
\begin{itemize}
% [topsep=0pt,itemsep=-1ex,partopsep=1ex,parsep=1ex]
    \item Section~\ref{additional_method} - Additional Method Details
    \item Section~\ref{exper_setup_details} - Experimental Setup \& Implementation Details
    \begin{itemize}
        \item Section~\ref{details_down_tasks} - Additional Implementation Details for Downstream Tasks
        \item Section~\ref{dataset_appendix} - Dataset Details
        \item 
        Section~\ref{prompt_templates} - Prompt Templates for Zero-Shot Evaluation
    \end{itemize}
    \item Section~\ref{comput_eff} - Computational Efficiency
    \item Section~\ref{add_ablation_studies} - Additional Ablation Studies
    \item
    Section~\ref{qual_results} - Qualitative Results
\end{itemize}

\section{Additional Method Details}
\label{additional_method}
\subsection{Formulation of GAT and GATv2}
\label{gat_gatv2_formula}
In Table \ref{tab:tab_6}, we investigate the effect of using attention maps as edge weights and compare our approach with alternative designs that learn attention coefficients from scratch, adopting the attention formulations of GAT \cite{velivckovic2018graph} and GATv2 \cite{brodyattentive}. In this section, we present the detailed formulations used to compute attention scores in GAT and GATv2. For GAT, the attention scores are computed using the following equation:

\begin{equation}
\alpha_{ij} =
\frac{
    \exp\!\left(
        \operatorname{LeakyReLU}\!\left(
            \vec{\mathbf{a}}^{T}
            \bigl[
                \mathbf{W} v_{i} \,\|\, \mathbf{W} v_{j}
            \bigr]
        \right)
    \right)
}{
    \sum_{k}
    \exp\!\left(
        \operatorname{LeakyReLU}\!\left(
            \vec{\mathbf{a}}^{T}
            \bigl[
                \mathbf{W} v_{i} \,\|\, \mathbf{W} v_{k}
            \bigr]
        \right)
    \right)
}
\end{equation}

\noindent where $\vec{\mathbf{a}}^{T} \in \mathbb{R}^{d \times d}$ denotes a learnable weight vector, and $\,\|\,$ denotes the concatenation operation. The $\operatorname{LeakyReLU}$ activation has a negative slope of 0.2. GATv2 demonstrates that GAT has limited expressive power for attention computation and reformulates the attention mechanism by modifying the order of operations to enhance this expressiveness. The corresponding equation is given by:

\begin{equation}
\alpha_{ij} =
\frac{
    \exp\!\left(
        \vec{\mathbf{a}}^{T}\!\left(
            \operatorname{LeakyReLU} \left(
                \mathbf{W}  \bigl[v_{i} \,\|\ v_{j}
            \bigr]\right)
        \right)
    \right)
}{
    \sum_{k}
    \exp\!\left(
        \vec{\mathbf{a}}^{T}\!\left(
            \operatorname{LeakyReLU} \left(
                \mathbf{W}  \bigl[v_{i} \,\|\ v_{k}
            \bigr]\right)
        \right)
    \right)
}
\end{equation}

\section{Experimental Setup \& Implementation Details}
\label{exper_setup_details}
\subsection{Additional Implementation Details for Downstream Tasks}
\label{details_down_tasks}
\noindent \textbf{Semantic Segmentation.} To enhance model robustness, we augment the training data with horizontal flipping, vertical flipping, and random rotations. We utilize the AdamW optimizer \cite{loshchilovdecoupled} with a learning rate of $1\mathrm{e}{-3}$ and weight decay of $1\mathrm{e}{-2}$. To mitigate class imbalance, as diseases typically occupy only a small portion of each image, we employ a linear combination of cross-entropy and Tversky losses. For the UPerNet \cite{xiao2018unified} decoder, feature maps are extracted from the frozen image encoder at layers \{2, 4, 7, 9, 11\} (zero-indexed). Training is performed on a single NVIDIA RTX 4090 GPU with a batch size of 4.

\vspace{3pt} 

\noindent \textbf{Object Detection.} We utilize the AdamW optimizer \cite{loshchilovdecoupled} with a learning rate of $1\mathrm{e}{-4}$ and weight decay of $1\mathrm{e}{-2}$. A linear warmup is applied during the first epoch, followed by a polynomial learning rate decay. For the YOLOv3 \cite{redmon2018yolov3} detection head, feature maps are extracted from the frozen image encoder at layers \{3, 7, 11\} (zero-indexed). The model is trained on a single NVIDIA RTX 4090 GPU with a batch size of 8.

\phantomsection
\subsection{Dataset Details}
\label{dataset_appendix}
\textbf{Histopathology Dataset Curation.} To construct a robust and domain-specific subset for our experiments, we curated a histopathology-focused dataset from the large-scale PMC-OA \cite{lin2023pmc} corpus. Given the multimodal nature of PMC-OA, which encompasses a broad range of medical imaging modalities, we applied a targeted keyword-based filtering protocol to isolate relevant histopathology samples. Specifically, we scanned the image captions across the entire dataset for a predefined set of diagnostic and technical terms. Samples were retained if their captions contained at least one of the following keywords: \texttt{h\&e, hematoxylin, eosin, histopathology, biopsy, or microscopic}. This automated filtering procedure ensures that the resulting subset primarily comprises tissue slides and microscopic structures, effectively excluding unrelated modalities such as radiological imaging.

\begin{table}[t]
\centering
% Adjusting column separation to fit 6 columns in narrow LNCS margins
\setlength{\tabcolsep}{4.5pt} 
\small 
\begin{tabular}{@{}l c c c c @{}}
\toprule
\multicolumn{1}{c}{\textbf{Task}} &  
\multicolumn{1}{c}{\textbf{Dataset}} & 
\multicolumn{1}{c}{\textbf{Classes}} & 
\multicolumn{1}{c}{\textbf{Total Size}} & 
\multicolumn{1}{c}{\textbf{Train/Val/Test (\%)}} \\ \midrule

Classification & CheXpert 5$\times$200 \cite{irvin2019chexpert} & 5 & 1000 & -/-/100 \\ \addlinespace
\midrule
Classification & RSNA \cite{shih2019augmenting} & 2 & 29,684 & -/--/15 \\ \addlinespace
\midrule
Classification &  SIIM \cite{SIIM-ACR} & 2 & 12,047 & --/--/30 \\ \addlinespace
\midrule
Classification & NIH ChestX-ray \cite{wang2017chestx} & 14 & 112,120 & --/5/- \\ \addlinespace
\midrule
Classification  & LC25000 (Lung) \cite{borkowski2019lung} & 3 & 15,000 & --/--/100 \\ \addlinespace
\midrule
Classification & LC25000 (Colon) \cite{borkowski2019lung} & 2 & 10,000 & --/--/100 \\ \addlinespace
\midrule
Classification & MHIST \cite{wei2021petri} & 2 & 3,152 & --/--/100 \\ \addlinespace
\midrule
Detection & RSNA \cite{shih2019augmenting} & 2 & 29,684 & 70/15/15 \\ \addlinespace
\midrule
Segmentation & SIIM \cite{SIIM-ACR} & 2 & 12,047 & 70/15/15 \\ 
\bottomrule
\end{tabular}
\caption{\textbf{Summary of dataset specifications and statistics.} ``Classes'' and ``Total Size'' refer to the number of categories and total images, respectively.}
\label{tab:dataset}
\end{table}

\noindent \textbf{Dataset Specifications and Statistics.} Table \ref{tab:dataset} details the specific characteristics of each downstream dataset used in evaluation, including the number of classes, image counts, and the respective training, validation, and test partitions. Additionally, Table \ref{tab:classes} lists the class names for each dataset.

\subsection{Prompt Templates for Zero-Shot Evaluation}
\label{prompt_templates}
Following CLIP \cite{radford2021learning}, we leverage simple prompt templates to align text and image features. For chest X-ray datasets, we utilize two templates: \texttt{"a chest X-ray image of \{disease\}"} and \texttt{"Findings suggesting \{disease\}"}. To ensure a fair comparison in the histopathology domain, we adopt the prompt templates from QuiltNet \cite{ikezogwo2023quilt}. Specifically, we use the following templates for histopathology datasets: \texttt{"a histopathology slide showing \{disease\}"}, \texttt{"histopathology image of \{disease\}"}, \texttt{"pathology tissue showing \{disease\}"}, and \texttt{"pres\-ence of \{disease\} tissue on image"}.

\begin{table}[t]
\centering
\small 
\setlength{\tabcolsep}{10pt} 
\begin{tabularx}{\textwidth}{@{}l >{\RaggedRight\arraybackslash}X @{}}
\toprule
% \multicolumn{number_of_columns}{alignment}{text}
\multicolumn{1}{c}{\textbf{Dataset}} & \multicolumn{1}{c}{\textbf{Classes}} \\ \midrule

CheXpert 5$\times$200 \mbox{\cite{irvin2019chexpert}} & "Atelectasis", "Cardiomegaly", "Consolidation", "Edema", "Pleural Effusion"\\ \addlinespace
\midrule
RSNA \mbox{\cite{shih2019augmenting}} & "No Finding", "Pneumonia" \\ \addlinespace
\midrule
SIIM \mbox{\cite{SIIM-ACR}} & "No Finding", "Pneumothorax" \\ \addlinespace
\midrule
NIH ChestX-ray \mbox{\cite{wang2017chestx}} & "Atelectasis", "Cardiomegaly", "Effusion", "Infiltration", "Mass", '"Nodule", "Pneumonia",
"Pneumothorax", "Consolidation", "Edema", "Emphysema", "Fibrosis", "Pleural Thickening", "Hernia" \\ \addlinespace
\midrule
LC25000 (Lung) \mbox{\cite{borkowski2019lung}} & "Lung adenocarcinoma", "Benign lung", "Lung squamous cell carcinoma" \\ \addlinespace
\midrule
LC25000 (Colon) \mbox{\cite{borkowski2019lung}} & "Colon adenocarcinoma", "Benign colonic tissue" \\ \addlinespace
\midrule
MHIST \mbox{\cite{wei2021petri}} & "Hyperplastic polyp", "Sessile serrated adenoma" \\ \addlinespace
\bottomrule
\end{tabularx}
\vspace{8pt}
\caption{\textbf{Class names for each dataset in zero-shot image classification.}}
\label{tab:classes}
\end{table}

\section{Computational Efficiency}
\label{comput_eff}
\begin{figure}
    \centering
    \includegraphics[width=0.65\linewidth]{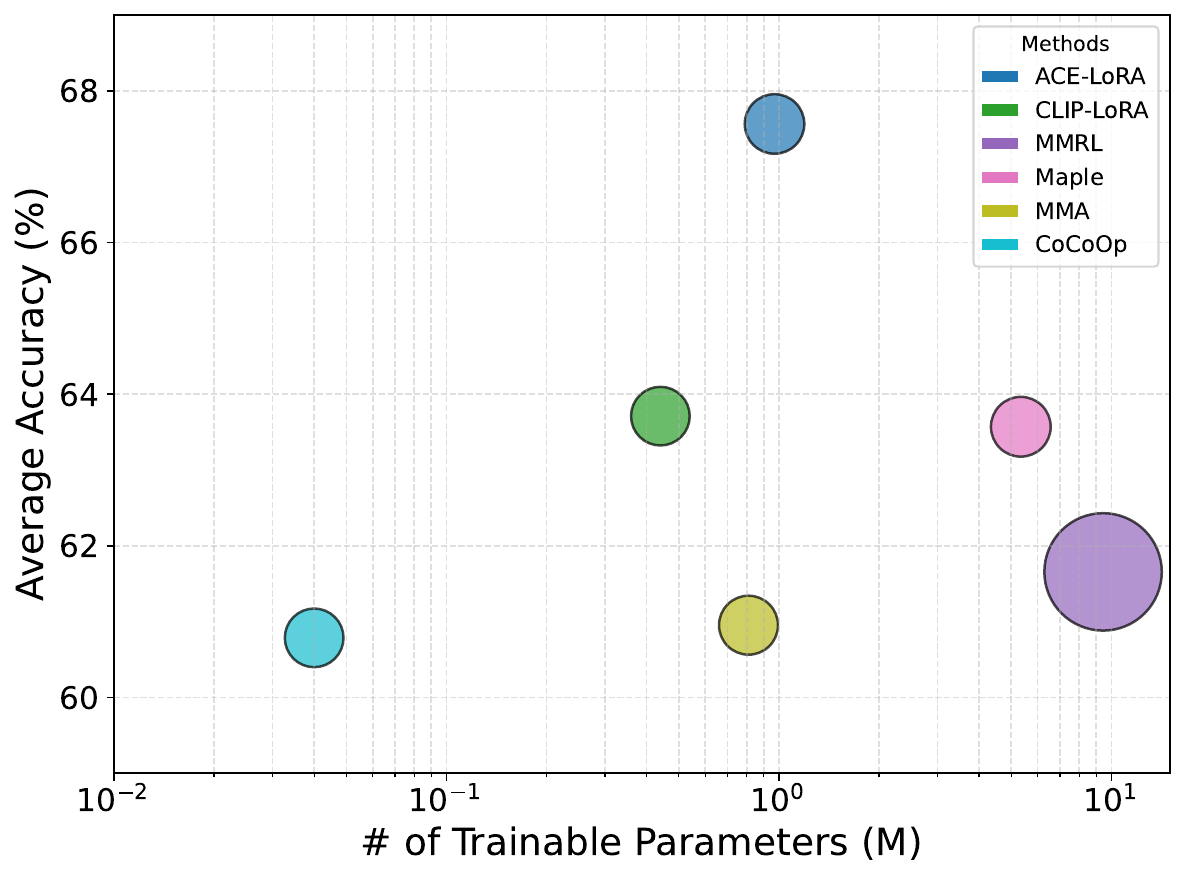}
    \caption{\textbf{Average zero-shot accuracy across three CXR benchmarks vs. number of trainable parameters (log-scale).} The bubble size denotes the computational cost in GFLOPs during the forward pass.}
    \label{fig_comp}

\end{figure}

We benchmark various PEFT methods against our approach, evaluating the trade-off between trainable parameters, forward-pass computational cost, and zero-shot accuracy averaged across the CheXpert 5$\times$200, RSNA, and SIIM data\-sets (Fig. \ref{fig_comp}). While CoCoOp is the most parameter-efficient, its performance remains suboptimal. In contrast, our approach achieves state-of-the-art performance with competitive parameter counts and computational overhead. Notably, our ACE-HGNN module incurs a marginal overhead over the LoRA baseline while yielding significant gains in accuracy.

\section{Additional Ablation Studies}
\label{add_ablation_studies}
\noindent \textbf{Effect of ACE-HGNN module integration on image and text encoders.} Table \ref{tab:tab_5} demonstrates the impact of applying the ACE-HGNN module to the image and text encoders in the zero-shot classification task. Notably, applying ACE-HGNN to the image encoder alone yields substantial gains across all datasets compared to the LoRA-only baseline. Similarly, applying the module solely to the text encoder also boosts performance, even achieving the best result on the CheXpert 5$\times$200 dataset. Overall, incorporating ACE-HGNN into either encoder consistently outperforms the LoRA-only baseline across all three benchmarks, underscoring the module's effectiveness. However, we observe that multimodal integration (i.e., applying the module to both encoders) delivers the most balanced results across the datasets.

\begin{table}[t]
  \centering
  \begin{tabular}{cc ccc}
    \toprule
     IE & TE & \textbf{ CheXpert } & \textbf{ RSNA } & \textbf{ SIIM } \\
    \midrule
     \ding{55} & \ding{55} & 45.80    & 72.85  &  72.50 \\
     \ding{51} & \ding{55} & 49.00 & 75.84 & \textbf{73.60} \\
     \ding{55} & \ding{51} & \textbf{50.40} & \underline{76.92} & 72.80 \\
     \rowcolor{SeaGreen!70} \ding{51} & \ding{51} & \underline{49.80} & \textbf{79.54} & \underline{73.35} \\
    \bottomrule
  \end{tabular}
  \caption{\textbf{Effects of ACE-HGNN module integration on the image encoder (IE) and the text encoder (TE).} Classification accuracy is reported in all datasets.}
  \label{tab:tab_5}
\end{table}

\noindent \textbf{Analysis of transductive inference.} To evaluate alternative approaches, we compare our method with TransCLIP \cite{zanella2024boosting}, a transductive approach that leverages unlabeled target-domain samples during inference, as shown in Table \ref{tab:tab_transclip}. While TransCLIP improves the performance of the baseline BiomedCLIP on two benchmarks, the gains are marginal and it even results in slightly worse performance on the RSNA dataset. These results suggest that unsupervised domain adaptation alone may be insufficient, motivating the need for domain adaptation with fine-tuning.

\begin{figure}
    \centering
    \includegraphics[width=0.65\linewidth]{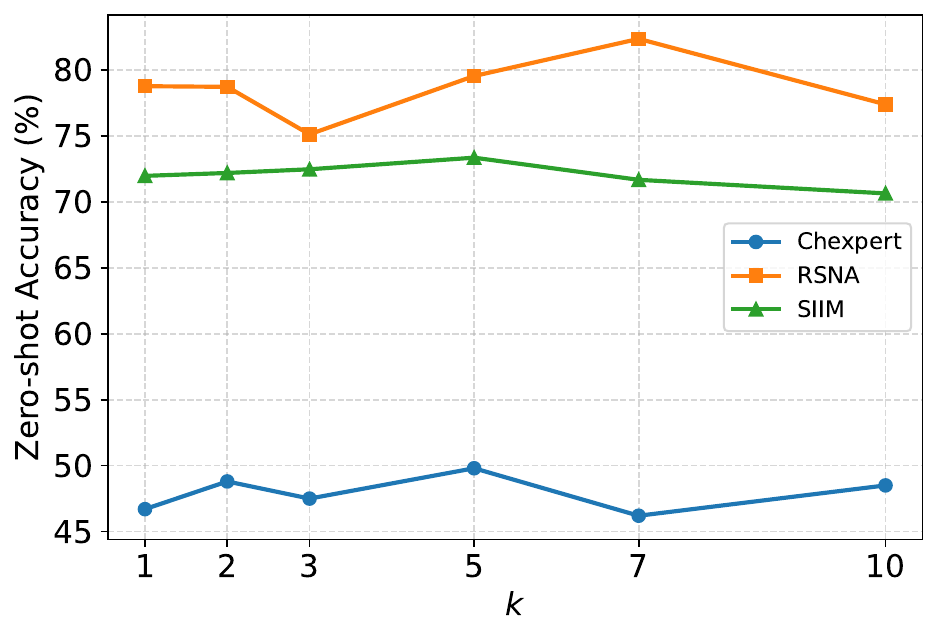}
    \caption{\textbf{Zero-shot accuracy across three datasets for varying $k$ values.} We find that selecting $k=5$ yields optimal performance.}
    \label{fig_topk}
\end{figure}

\noindent \textbf{Effect of different $k$ values in the top-$k$ operation.} We conduct an ablation study to investigate the impact of the number of selected neighbor tokens on performance, as illustrated in Figure \ref{fig_topk}. Notably, even with $k = 1$, ACE-HGNN achieves substantial zero-shot gains over the baseline fine-tuned solely with LoRA (see Table \ref{tab:tab_4}). While $k = 5$ yields optimal results, performance remains consistently high and robust across all tested $k$ values and benchmarks, highlighting the effectiveness of the ACE-LoRA framework.

\begin{figure}
    \centering
    \includegraphics[width=0.65\linewidth]{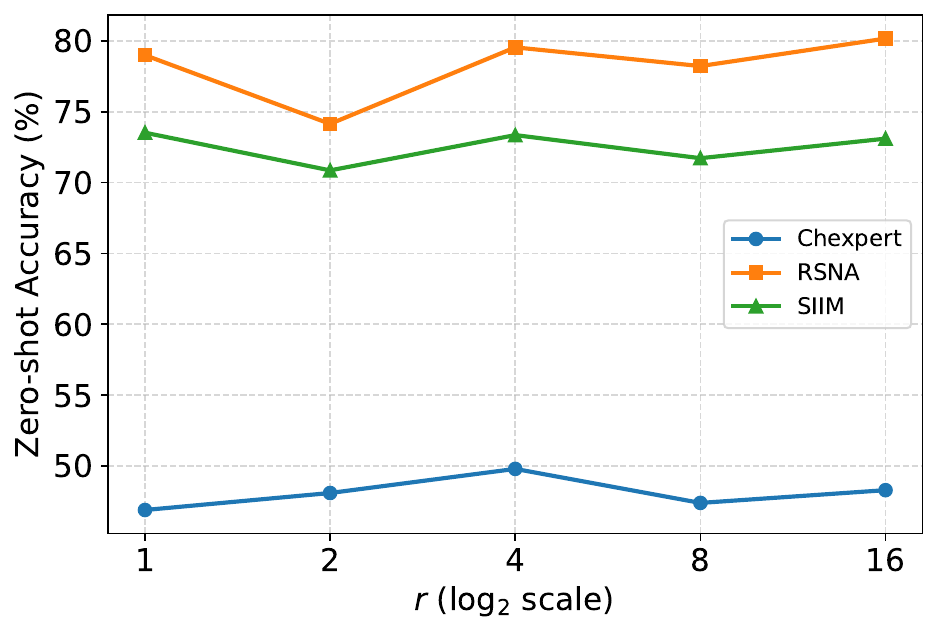}
    \caption{\textbf{Zero-shot accuracy across three datasets for different $r$ values.} Setting the rank $r = 4$ delivers the best performance.}
    \label{fig_rank}
\end{figure}

\noindent \textbf{Impact of different rank $r$ values.} To assess how the rank of the integrated LoRA modules affects performance, we vary the $r$ values as shown in Figure \ref{fig_rank}. We find that ranks below 4 yield weaker performance compared to $r=4$, suggesting that such low-rank adaptations lack sufficient representational complexity. In contrast, increasing the rank beyond 4 provides no additional gains and even slightly degrades accuracy.

\begin{table}[t]
  \centering
  \begin{tabular}{l ccc}
    \toprule
     \textbf{Method} & \textbf{ CheXpert } & \textbf{ RSNA } & \textbf{ SIIM } \\
    \midrule
     BiomedCLIP \cite{zhang2023biomedclip} & 35.50 & \textbf{74.34} & 61.40 \\
     \rowcolor{SeaGreen!70} TransCLIP \cite{zanella2024boosting} & \textbf{37.50} & 73.54 & \textbf{65.36} \\
    \bottomrule
  \end{tabular}
  \caption{\textbf{Analysis of transductive inference.} Classification accuracy is reported in all datasets.}
  \label{tab:tab_transclip}
\end{table}

\begin{figure}
    \centering
    \includegraphics[width=0.65\linewidth]{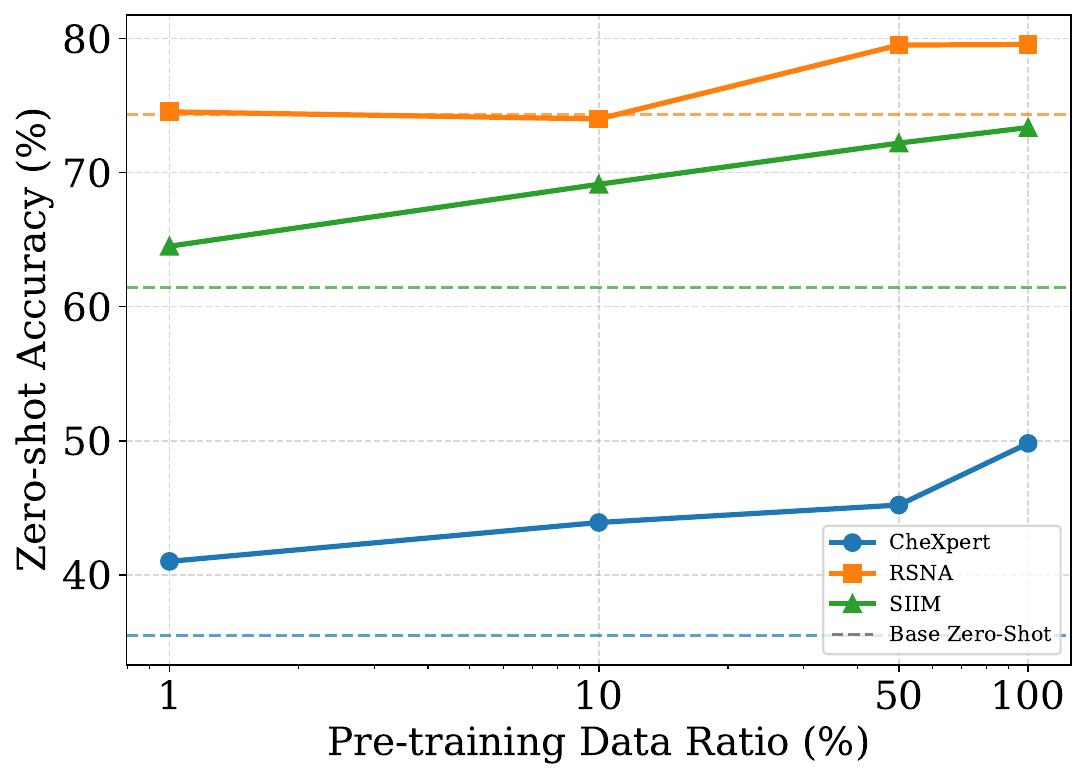}
    \caption{\textbf{Zero-shot accuracy as a function of pre-training data scale (log-scale).} Dashed lines indicate the zero-shot performance of BiomedCLIP without domain adaptation.}
    \label{fig_scaling}
\end{figure}

\noindent \textbf{Impact of pre-training data scale.} To characterize the framework's scaling behavior, we evaluate performance across a data regime ranging from $1\%$ to $100\%$ of the pre-training set. This analysis investigates how effectively the model leverages increasing data volumes to learn transferable representations. As summarized in Figure \ref{fig_scaling}, we observe a consistent positive correlation between pre-training data scale and downstream performance. While even a minimal amount of data ($1\%$) improves over the baseline, performance continues to increase as the data scale approaches $100\%$. These results demonstrate that our framework scales effectively with larger datasets, consistently improving representation quality and yielding superior performance on downstream benchmarks.

\vspace{-1cm}

%\noindent \textbf{t-SNE visualization of image embeddings.} Figure \ref{fig_5} presents a comparison of t-SNE \cite{maaten2008visualizing} visualizations of image embeddings from BiomedCLIP and ACE-LoRA. For BiomedCLIP, embeddings are extracted from the final layer of the image encoder, whereas for ACE-LoRA, we utilize the refined image embeddings produced by the ACE-GNN module for visualization. The results demonstrate that our fine-tuning framework exhibits more segregated class-representations than the base model, helping explain its superior classification performance.

%\begin{figure}
%    \centering
%    \includegraphics[width=\linewidth]{figs/tsne.pdf}
%    \caption{\textbf{t-SNE visualizations of image embeddings on CheXpert 5$\times$200.} Each color corresponds to a different disease.}
%    \label{fig_5}
%\end{figure}

\section{Qualitative Results}
\label{qual_results}
\begin{figure*}
    \centering
    \includegraphics[width=\linewidth]{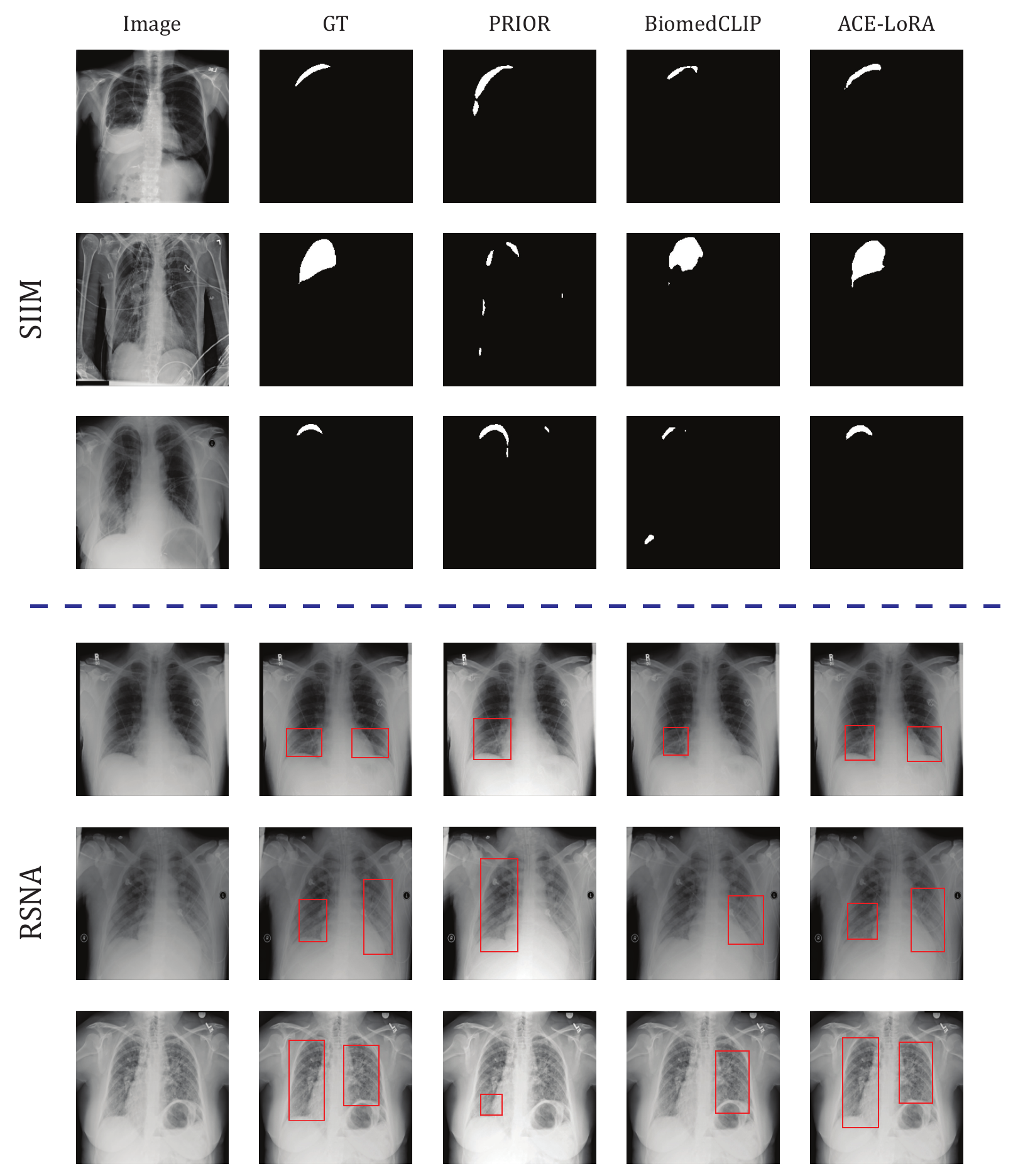}
    \caption{\textbf{Qualitative comparison of different image encoder backbones.} We compare our method with PRIOR \cite{cheng2023prior} and BiomedCLIP \cite{zhang2023biomedclip} on the semantic segmentation and object detection tasks. GT denotes the ground truth.}
    \label{fig:appendix}
\end{figure*}

Figure \ref{fig:appendix} illustrates segmentation and detection predictions across different image encoder backbones on the SIIM \cite{SIIM-ACR} and RSNA \cite{shih2019augmenting} datasets, respectively. ACE-LoRA outperforms established medical VLMs, PRIOR \cite{cheng2023prior} and BiomedCLIP \cite{zhang2023biomedclip}, on both downstream tasks.

\end{document}